# HausaMovieReview: A Benchmark Dataset for Sentiment Analysis in Low-Resource African Language

Asiya Ibrahim Zanga[1], Salisu Mamman Abdulrahman[2], Abubakar Ado[3], Abdulkadir Abubakar Bichi[3], Lukman Aliyu Jibril[4], Abdulmajid Babangida Umar[3], Alhassan Adamu[2] Shamsuddeen Hassan Muhammad[5], Bashir Salisu Abubakar

[1]Department of Computer Science, Federal University Dutsin-Ma, Katsina, Nigeria (email: azanga@fudutsinma.edu.ng).
[2]Department of Computer Science, Aliko Dangote University of Science and Technology, Wudil, Kano, Nigeria (emails: salisu.abdul@kustwudil.edu.ng; alhassanadamu@kustwudil.edu.ng, bsalisu@gmail.com).
[3]Department of Computer Science, Northwest University, Kano, Nigeria (emails: aarogo@yumsuk.edu.ng, aabichi@yumsuk.edu.ng, abumar@yumsuk.edu.ng;).
[4](email: lukman.j.aliyu@gmail.com).
[5]Department of Computer Science, Bayero University Kano, Nigeria (email: shamsudden2004@gmail.com).

**Abstract**

The development of Natural Language Processing (NLP) tools for low-resource languages is critically hindered by the scarcity of annotated datasets. This paper addresses this fundamental challenge by introducing HausaMovieReview, a novel benchmark dataset comprising 5,000 YouTube comments in Hausa and code-switched English. The dataset was meticulously annotated by three independent annotators, demonstrating a robust agreement with a Fleiss' Kappa score of 0.85 between annotators. We used this dataset to conduct a comparative analysis of classical models (Logistic Regression, Decision Tree, K-Nearest Neighbors) and fine-tuned transformer models (BERT and RoBERTa). Our results reveal a key finding: the Decision Tree classifier, with an accuracy and F1-score 89.72% and 89.60% respectively, significantly outperformed the deep learning models. Our findings also provide a robust baseline, demonstrating that effective feature engineering can enable classical models to achieve state-of-the-art performance in low-resource contexts, thereby laying a solid foundation for future research.

## 1. Introduction

Nigeria has over 500 languages, but lack of datasets has led to their underrepresentation or exclusion from natural language processing (NLP) studies[1]. The amount of unstructured text has increased dramatically due to the internet's spectacular improvement. The proliferation of smart mobile devices and the innumerable number of mobile applications finding an analytical procedure that enables analysts to retain text in a way that will help them comprehend human ideas was therefore becoming more and more necessary. Being subjective beings, humans are capable of complex emotional expression. However, any choice they choose will be influenced by their opinions, it becomes vital to train information systems to identify persons at that level in order to be able to comprehend these viewpoints from vast volumes of material, which calls for more work from researchers[2].

While some progress has been made in bridging this gap like [3] many benchmark datasets for African languages are limited to a single domain, making them difficult to apply to other areas of interest [4]Sentiment analysis is one of the most popular tasks in NLP, with datasets for high-resource languages like English available across different domains, including Twitter, social media updates, product reviews and movie critiques. It is the task of identifying the emotional tone of text and has become a cornerstone of various fields, including marketing, customer service and social media monitoring[5]. While significant advancements have been made in sentiment analysis for widely spoken languages, research on low-resource languages like Hausa remains relatively under-explored.

However, for movie review in Nigeria, the only available dataset is NollySenti [4], a Twitter sentiment classification dataset for five languages including English, Hausa, Igbo, Yoruba and Nigerian-pidgin. Despite the approach contributing significantly to the domain of movie review in these Nigerian languages, its applicability to other domains remains unclear.

The Hausa-language cinema industry known as Kannywood, which is based in Northern Nigeria, has a large following and a profound cultural influence throughout the Sahel region of Africa. Producers, directors and marketers can help shape their tactics by gaining useful insights into audience preferences and opinions through sentiment analysis of reviews and social media comments regarding Kannywood movies. Nevertheless, compared to many other languages, the field of natural language processing (NLP) for Hausa is less established, which makes it difficult to extract sentiment from writings written in the language. The Kannywood industry has witnessed a surge in popularity, yet analyzing the sentiment expressed in these movies remains a challenge due to the lack of language-specific sentiment analysis tools. Existing sentiment analysis models often rely on large datasets, which are scarce for Hausa[6] Furthermore, the unique linguistic characteristics of Hausa, including its complex morphology and dialectal variations, pose additional hurdles for accurate sentiment classification.

This study aims to address this gap by developing a sentiment analysis model for Kannywood movies, a popular Nigerian Hausa-language film industry, we propose a method to aid in the collection, filtering and annotation of data for a low-resource language.

We introduce HausaMovieReview, an open-source sentiment dataset based on reviews of Hausa-language movies. Using comments from the LABARINA series, we collected 17,095 YouTube comments and manually annotated 5000 to build the Kannywood corpus for sentiment classification named HausaMovieReview for Hausa-language. Kannywood, which produces films about the Kano people, highlights Nigeria's diverse cultures with emphasis on the northern culture and is one of the country's largest film industries. the research contributions are summarized as follows:
  i. Datasets creation
  ii. Building a fine-tuned transformer model (BERT and RoBERTa),
  iii. Perform a comparative analysis using classical Machine Learning and Deep Learning models.

## 2. Literature Review

Sentiment analysis, also known as opinion mining, involves processing textual data to assess and interpret the emotions expressed within it. Understanding "sentiment" is crucial; while often equated with feelings, sentiments focus more on opinions and attitudes than on factual information. According [7], feelings are instinctive responses to stimuli (e.g., joy, anger), while sentiment is an emotion influenced by one's opinions or perceptions. Psychologists categorize feelings into six primary emotions: love, joy, surprise, anger, sadness and fear. Sentiment analysis primarily aims to determine whether an opinion is positive, negative, or neutral, thereby simplifying complex emotional scenarios.

One key aspect of sentiment is its universality; everyone experiences emotions, making each person a potential source of valuable data for sentiment analysis tools. This intrinsic characteristic of human emotions makes sentiment analysis a powerful tool across various domains, including business, politics and social media. Sentiment analysis has become an increasingly important area of research in natural language processing. Previous studies have explored various approaches to sentiment classification, including rule-based methods, machine learning techniques and deep learning models. However, there is still a lack of research on sentiment analysis for Kannywood movie reviews. Sentiment analysis has evolved significantly over the years. Early approaches relied primarily on lexicon-based methods, which involved using predefined sentiment lexicons to classify text[8]. Though sentiment analysis has been the focus of study in education [9], it has also yielded numerous successful applications in other domains, including as business [10]and medicine [11]

Various machine learning algorithms were explored for sentiment analysis, including Support Vector Machines (SVM), Decision Trees (DT), Random Forests (RF), Multilayer Perceptron (MLP) and Long Short-Term Memory (LSTM) networks[12]. While lexicon-based approaches were also considered, machine learning algorithms generally outperformed them in terms of accuracy[13]. Several researchers have proposed cross-lingual NLP approaches to address the challenges of low-resource languages by leveraging resources from high-resource languages like English [14]. For sentiment analysis, a common strategy is to translate comments from the original low-resource language (e.g., Hausa) into English and then apply well-performing English sentiment analysis models.

Research on sentiment analysis in the movie domain has predominantly focused on English-language reviews. Early works by [15] established foundational approaches using machine learning models and manually engineered features. Subsequent studies explored deep learning techniques such as Convolutional Neural Networks (CNNs) and Recurrent Neural Networks (RNNs), which significantly improved sentiment prediction accuracy [16]. However, the advent of transformer-based architectures like BERT and RoBERTa has set a new benchmark, with studies reporting state-of-the-art results across various sentiment analysis tasks [17].

### 3. Methodology

Among the objectives of this study is to develop a HausaMovieReview dataset and to evaluate a sentiment analysis model capable of accurately classifying the Kannywood movie comments into three distinct sentiment categories: *positive*, *neutral* and negative. To achieve this, a comprehensive methodology was designed integrating both classical machine learning (ML) techniques and advanced transformer-based models. This chapter details the entire process including the dataset preparation, the implementation of each model, the experimental setup and the chosen evaluation metrics used to assess performance.

### 3.1 Dataset Construction

This section details the meticulous process undertaken to construct the HausaMovieReview dataset, a critical contribution of this research. The methodology encompasses data collection from a widely accessible online platform, a systematic sampling procedure, a rigorous three-phase annotation process by independent human experts and a comprehensive analysis of inter-annotator agreement.

**Data Collection and Sampling**

The primary source of data for this study is the comment sections of 13 episodes of the popular Kannywood series "Labarina" on YouTube. YouTube, as a global social media platform serves as a significant hub for user-generated content and provides a rich source of genuine, spontaneous opinions. A web scraper was developed and utilized to collect all comments from these specific episodes. This collection resulted in a total of 17,095 raw comments.

From this initial corpus, a subset of 5,000 comments was randomly sampled to form the final dataset for annotation. This sampling approach ensured that the annotated dataset was representative of the broader spectrum of opinions expressed on the platform, mitigating potential biases that could arise from selecting comments based on popularity or content. To maintain the integrity of the data, the comments were not pre-processed or altered in any way before annotation. The comments contain a mixture of pure Hausa, code-switched English-Hausa and occasional loanwords from Arabic, reflecting the typical linguistic patterns observed in online communication within the region.

### 3.2 Annotation Process

The annotation task was performed by three independent native Hausa speakers, each with a strong understanding of the language's nuances, idioms and cultural context. All three annotators possessed a deep familiarity with the Kannywood film industry, which was crucial for interpreting context-dependent comments. The annotation process was structured in three key phases:

1. **Label Definition and Training**: A clear set of guidelines was developed to define the three sentiment labels: Positive, Neutral and Negative.
   i. **Positive**: A comment expressing approval, praise, happiness, or a favorable opinion. Examples include "Masha Allah, a beautiful movie," or "Great acting!"
   ii. **Neutral:** A comment that does not convey a clear sentiment. This includes factual statements, questions, greetings, or off-topic remarks. Examples include "When is the next episode?" or "Hello from Niger."
   iii. **Negative:** A comment expressing disapproval, criticism, sadness, or a negative opinion. Examples include "This movie is a waste of time" or "The plot is very bad." The annotators were trained on these guidelines using a small pilot set of comments to ensure a shared understanding and consistent application of the labeling criteria.

2. **Independent Annotation:** Each of the three annotators independently labeled the entire set of 5,000 comments according to the defined guidelines. This independent approach was essential for establishing a reliable ground truth and for subsequently measuring the degree of agreement between them.

3. **Majority Vote and Finalization**: Upon completion of the independent annotations, the final label for each comment was determined using a majority-vote mechanism. If two or more annotators agreed on a sentiment, that sentiment was assigned as the final label. In cases where all three annotators assigned a different label (a rare occurrence), the comment was flagged for review and a consensus was reached through discussion. This process ensured that the final dataset, referred to as HausaMovieReview, was a highly reliable and consistent resource.

### 3.3 Inter-Annotator Agreement

To validate the reliability and consistency of the annotation process, two key metrics were used to measure inter-annotator agreement (IAA): Cohen's Kappa (κ) and Fleiss' Kappa (κ). Cohen's Kappa measures pairwise agreement between two annotators, while Fleiss' Kappa extends this to measure the agreement among all three annotators. The scores obtained were provided in Table 1 and Table 2 as described below.

Table 1: Inter-Annotator Agreement Scores

| Pair | Cohen's/Fleiss' Kappa |
|---|---|
| Annotator 1 vs Annotator 2 | 0.7975 |
| Annotator 1 vs Annotator 3 | 0.9071 |
| Annotator 2 vs Annotator 3 | 0.8903 |
| Fleiss' Kappa (All Annotators) | 0.865 |

Table 2: Annotator Label Distributions

| Annotator | Negative | Neutral | Positive |
|---|---|---|---|
| Annotator 1 | 1165 (23.30%) | 996 (19.92%) | 2839 (56.80%) |
| Annotator 2 | 1165 (23.30%) | 995 (19.90%) | 2840 (56.80%) |
| Annotator 3 | 1165 (23.30%) | 995 (19.90%) | 2840 (56.80%) |
| Majority Vote | 1165 (23.30%) | 995 (19.90%) | 2840 (56.80%) |

These high IAA scores are a testament to the quality of the dataset and the clarity of the annotation guidelines, ensuring that the HausaMovieReview dataset is a robust and trustworthy resource for sentiment analysis research. The complete dataset and all associated code are publicly available on GitHub.
https://github.com/AsiyaZanga/HausaMovieReview.git

### 3.4 Experiment Setup

This section details the experiment design used for this study including experiment steps, classical machine learning and deep learning and training settings

**Preprocessing**

The dataset for this study was a corpus of 5,000 comments extracted from comment sections of 13 episodes of LABARINA. Given the informal and multi-lingual nature of the data, a rigorous preprocessing pipeline was established. This involved converting all text to lowercase, removing punctuation, special characters and tokenizing the sentences into individual words.

**Feature Extraction: Term Frequency-Inverse Document Frequency (TF-IDF)**

The TF-IDF vectorization technique was employed to transform the preprocessed text data into a matrix of numerical features. TF-IDF assigns a weight to each word in a document, reflecting its importance within that specific comment relative to the entire corpus. A high TF-IDF score indicates a term that is both frequent in a document and rare across the entire dataset, making it a powerful feature for distinguishing between sentiment classes.

**Model Implementation**
Following the feature extraction, three distinct classical ML classifiers were trained and evaluated:

1. **Logistic Regression (LR):** This is a linear classification model that uses a logistic function to estimate the probability of a given comment belonging to a particular sentiment class. Despite its simplicity, LR is highly effective for text classification tasks, providing a strong baseline for comparison.
2. **Decision Tree (DT):** A non-linear, supervised learning model that partitions the data into a series of decisions, creating a tree-like structure. The DT model makes predictions by traversing the tree from the root to a leaf node, where each internal node represents a feature test and each leaf node represents a class label.
3. **K-Nearest Neighbors (KNN):** KNN is an instance-based, non-parametric algorithm that classifies a new data point based on the majority class among its k nearest neighbors in the feature space. The value of k was a key hyperparameter tuned during the training process to optimize performance.

**Transformer-based Approach**

To capture the complex semantic and contextual nuances of the comments, pre-trained transformer models were leveraged. This approach bypassed the need for manual feature engineering by using the models' inherent ability to learn rich, contextual embeddings from raw text. We have selected to models as described below:

1. **BERT (Bidirectional Encoder Representations from Transformers):** The BERT-base-uncased model was selected for its ability to learn bidirectional representations from a massive corpus of text. The model was fine-tuned for the sentiment classification task by adding a simple classification layer on top of its output, allowing it to adapt its pre-trained knowledge to the specific domain of Kannywood comments.
2. **RoBERTa (A Robustly Optimized BERT Pretraining Approach):** RoBERTa, an enhanced version of BERT, was also chosen. It was trained with an optimized methodology, including a larger dataset, a longer training period and the removal of the next sentence prediction objective. This robust training process often results in superior performance on downstream tasks compared to standard BERT.

**Training Setup**

The fine-tuning of both BERT and RoBERTa was conducted using a standard training protocol. The final classification layer was added to the pre-trained models. Key training parameters were carefully configured: a small learning rate $2\times10^{-5}$, a batch size of 16 and a limited number of 3 epochs to prevent overfitting. The AdamW optimizer was used, as it is widely recommended for training transformer models.

### 3.5 Evaluation Method and Metrics

To ensure a fair and robust evaluation of all models, a consistent experimental protocol was followed. The dataset was not partitioned into separate training and testing sets. Instead, a crucial aspect of the training protocol was the use of 10-fold cross-validation. This technique partitioned the data into ten subsets, training the model on nine and validating on the tenth, repeating this process ten times. This method ensured that the model's performance was not dependent on a specific data split and reduced the risk of overfitting.

Model performance was assessed using a suite of standard metrics for multi-class classification:

i. **Accuracy:** The ratio of correctly predicted instances to the total number of instances.
ii. **Precision:** The ratio of true positive predictions to the total positive predictions made by the model. It indicates the model's ability to avoid false positives.
iii. **Recall:** The ratio of true positive predictions to all actual positive instances. It measures the model's ability to find all relevant instances.
iv. **F1-Score:** The harmonic mean of precision and recall. It provides a balanced measure of a model's performance, particularly useful for imbalanced datasets. The formula for the
v. **Area Under the Curve (AUC):** A measure of the model's ability to distinguish between classes. A higher AUC value indicates a better-performing model.

Accuracy, precision, recall, F1-score, and AUC are widely recognized evaluation metrics for classification tasks, as they capture complementary aspects of model performance[18]. Prior studies in meta-learning and algorithm selection confirm these measures as standard practice for assessing predictive effectiveness[19]

## 4. Results and Discussion

In this section, we present the comprehensive results obtained from the sentiment analysis experiments conducted on the HausaMovieReview dataset. It details the performance of both classical machine learning models (Logistic Regression, Decision Tree, K-Nearest Neighbors) and deep learning models (BERT and RoBERTa).

### 4.1 Model Performance Evaluation

The key performance metrics Accuracy, F1-Score, were computed for each model with Precision, Recall and AUC in addition for classical models. The results, including their corresponding standard deviations, are summarized in table 3 below.

Table3: Model Performance for Classical Models

| Model | Accuracy | Precision | Recall | F1 | AUC |
|---|---|---|---|---|---|
| Decision Tree | 89.71% | 90.02% | 89.71% | 89.60% | 93.45% |
| KNN | 63.36% | 76.89% | 63.36% | 62.11% | 86.27% |
| Logistic Regression | 86.81% | 87.08% | 86.81% | 86.81% | 95.92% |

The results in Table 1 indicate that all three classical models achieved reasonable to excellent performance. The Decision Tree classifier demonstrated superior performance across most metrics, achieving the highest mean accuracy (0.8971) and a strong F1-score (0.896). This suggests that a rule-based, non-linear approach was highly effective at classifying the sentiment of the Hausa comments. The Logistic Regression model also performed exceptionally well, with a high accuracy of 0.8681 and the highest AUC score of 0.9592, which indicates its excellent ability to distinguish between all three sentiment classes. In contrast, the KNN model exhibited the lowest performance, with a mean accuracy of 0.6336, suggesting it was less effective at handling the high-dimensional, sparse TF-IDF data. Below are graph for pictorial representations.

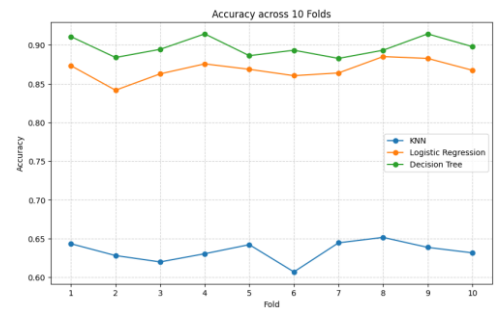

**Figure 1: Accuracy of KNN, Logistic Regression and Decision Tree Models Across 10 Folds**

Figure 1 illustrates the accuracy of the three models over 10 separate validation folds. The plot clearly shows that the Logistic Regression and Decision Tree models consistently perform better than the KNN model, with accuracy scores generally above 0.85 and 0.90, respectively. The KNN model's accuracy remains significantly lower, fluctuating between 0.60 and 0.65.

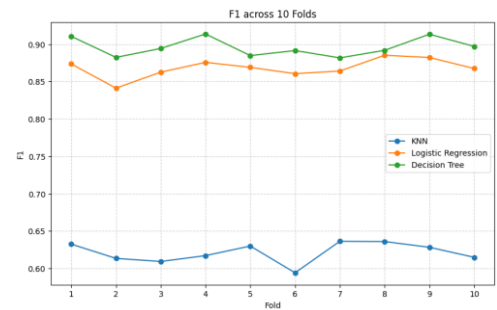

*Figure 2: F1 Score of KNN, Logistic Regression and Decision Tree Models Across 10 Folds*

Figure 2, displays the F1 score for each model across 10 folds. The plot demonstrates a similar pattern to the accuracy graph, with the Logistic Regression and Decision Tree models exhibiting much higher F1 scores than the KNN model.

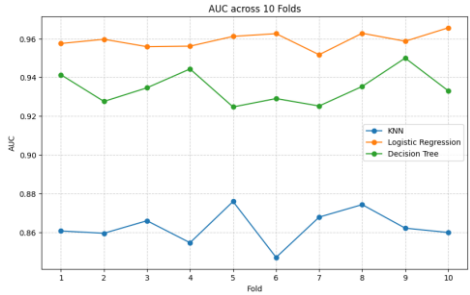

*Figure 3: AUC Score of KNN, Logistic Regression and Decision Tree Models Across 10 Folds*

Figure 3 shows the Area Under the Curve (AUC) score for each model over 10 folds. According to the plot, Logistic Regression consistently has the highest AUC score, often above 0.95, while the Decision Tree model's AUC is also high, typically above 0.92. The KNN model, in contrast, shows a much lower and more volatile AUC score.

**Deep Learning Model Results: BERT**
The BERT model was fine-tuned on the HausaMovieReview dataset to analyze its performance throughout the training process. The following sections present the training and validation performance, followed by the final evaluation on the test set..

**Comparative Results of Deep Learning Models**
Both the BERT and RoBERTa models were fine-tuned on the KannySenti dataset, demonstrating effective learning and similar performance trends. During training, both models showed a rapid increase in validation accuracy and F1-score in the early stages, followed by a plateauing of these metrics and a fluctuation in validation loss. This behavior, particularly after Step 40 for both models, suggests that they began to overfit the training data.
On the final held-out test set, the BERT model slightly outperformed RoBERTa, achieving an accuracy of 79.7% and an F1-score of 0.7562, compared to RoBERTa's 76.6% accuracy and 0.7292 F1-score. Table 4 below and subsequent graphs explain more.

**Table 4 Evaluation Result for Transformer Models**

| Model | Evaluation Loss | Evaluation Accuracy | Evaluation F1-Score |
|---|---|---|---|
| BERT | 0.5906 | 0.797 | 0.7562 |
| RoBERTa | 0.6855 | 0.766 | 0.7292 |

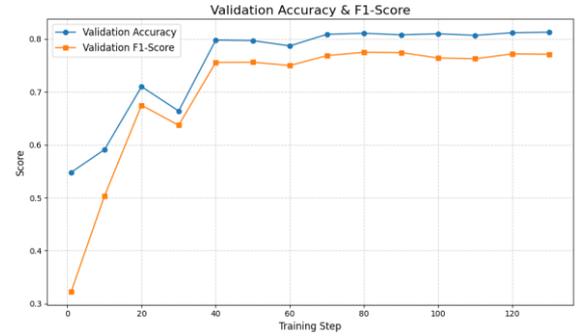

*Figure 4 Validation Accuracy and F1-Score during BERT Model Fine-tuning*

Figure 4 Shows the validation accuracy and macro-averaged F1-score of the BERT model against training steps. It visually represents the model's performance on unseen validation data, highlighting its convergence and predictive capability throughout the fine-tuning process.

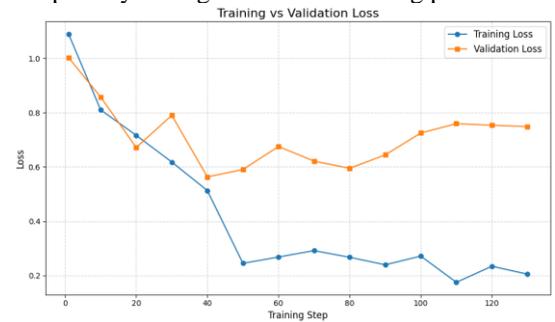

*Figure 5 Training and Validation Loss during BERT Model Fine-tuning*

Figure 5 illustrates the progression of training loss and validation loss for the BERT model across various training steps. It demonstrates the model's learning curve on the training data and its generalization behavior on the validation set, indicating potential points of overfitting.

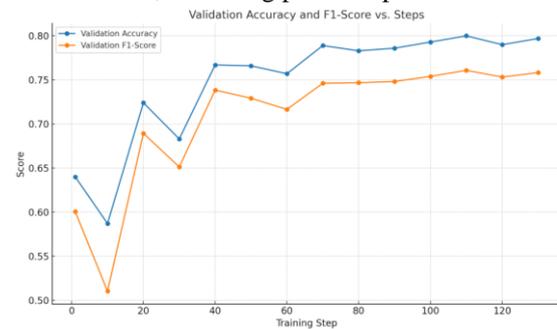

**Figure 6: Validation Accuracy and F1-Score during RoBERTa Model Fine-tuning**

Figure 6 Shows the validation accuracy and macro-averaged F1-score of the RoBERTa model against training steps. It visually represents the model's performance on unseen validation data, indicating convergence and overall predictive capability over the fine-tuning process.

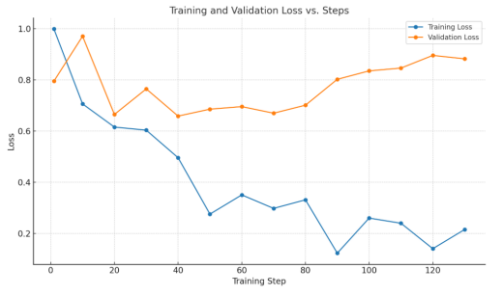

**Figure 7 Training and Validation Loss during RoBERTa Model Fine-tuning**

Figure 7 illustrates the progression of training loss and validation loss for the RoBERTa model across various training steps. A sustained decrease in training loss indicates effective learning on the training data, while the behavior of validation loss highlights the model's generalization capabilities and potential for overfitting.

## 5. Discussion of Findings

The most significant finding of this study is the remarkable performance of the **Decision Tree model**, which outperformed both the classical (Logistic Regression and KNN) and the more advanced transformer-based models, (BERT and RoBERTa). This result is counterintuitive, as transformer models are generally considered state-of-the-art for natural language processing tasks due to their ability to capture complex, contextual relationships in text.

Several factors may explain this surprising outcome. The dataset, consisting of Kannywood movie comments, is relatively small (5,000 comments). Transformer models, particularly large pre-trained ones like BERT and RoBERTa, require vast amounts of data to fully realize their predictive power. In a limited data environment, these models may be prone to overfitting or fail to fine-tune their extensive pre-trained knowledge effectively to the specific nuances of the Kannywood domain.

In contrast, the Decision Tree model, a non-linear classifier, proved exceptionally well-suited to the task. It appears to have effectively identified key features within the TF-IDF vectorized data that are highly predictive of sentiment, such as specific keywords, phrases, or their combinations. Its hierarchical, rule-based nature may have allowed it to create a series of effective decision rules that generalize well to the local data distribution, outperforming more complex models that could not fully leverage their potential on a small dataset. This finding suggests that for domain-specific and size-constrained text corpora, carefully engineered classical models can be more effective and computationally efficient than resource-intensive transformers.

**Dataset Reliability and Limitations**

The reliability of our dataset is a critical factor in the validity of these results. As detailed in the methodology, the dataset underwent a three-way human annotation process. The high Inter-Annotator Agreement (IAA), achieved through a majority voting protocol, confirms the consistency and quality of the sentiment labels. This high IAA score suggests that the sentiment labels are reliable and that the models were trained on a high-quality ground-truth dataset.

Despite these promising results, several limitations of this study must be acknowledged. First, the dataset is restricted to Kannywood movie comments and may not be generalizable to other domains or languages. The models' performance on a broader range of text would likely differ. Second, while the classical models outperformed the transformers in this specific experiment, this does not invalidate the general superiority of transformer architectures. It merely highlights the importance of data volume and domain-specificity for fine-tuning these models. A future study could explore pre-training a transformer model specifically on a massive corpus of Hausa-language text, which may then yield superior results on a Kannywood-specific task. Finally, the comments contained a significant amount of code-switching between Hausa and English, as well as informal language and slang, which poses a unique challenge for any model.

## 6. Conclusion

This research addressed the critical gap in sentiment analysis tools for Hausa-language movie reviews within the rapidly growing Kannywood film industry. By creating the HausaMovieReview dataset, a novel resource of 5,000 manually annotated YouTube comments from the LABARINA series, this study provided a crucial foundation for advancing NLP in this under-resourced domain.

The study investigated the application of both classical machine learning and deep learning models for sentiment classification. Surprisingly, the classical machine learning models, particularly the Decision Tree classifier, achieved exceptional performance on the HausaMovieReview dataset, with an accuracy of 89.71% and a macro-averaged F1-score of 90.02%. This indicates that for this specific dataset, TF-IDF features combined with a Decision Tree proved remarkably effective.

The fine-tuned deep learning models, BERT ('bert-base-uncased') and RoBERTa ('cardiffnlp/twitter-roberta-base-sentiment-latest'), also demonstrated strong capabilities, achieving accuracies of 79.7% and 76.6% and F1-scores of 75.62% and 72.92%, respectively. While these results are commendable, they were notably lower than those achieved by the classical Decision Tree model in this study.

## 7. Future Work

Building upon the findings and addressing the limitations of this study, several promising avenues for future research emerge:

- Dataset Expansion and Diversification: Future efforts should focus on significantly expanding the KannySenti dataset by annotating a larger and more diverse collection of Kannywood movie reviews from various online platforms and genres.
- Hausa-Specific NLP Resources and Preprocessing: Research into developing more robust Hausa-specific NLP tools (e.g., improved tokenizers, stemmers/lemmatizers) and preprocessing techniques tailored to Hausa linguistic features and code-switching patterns is crucial.
- Deeper Investigation into Classical ML Performance: Conduct further analysis to understand why classical models, particularly Decision Tree with TF-IDF, performed exceptionally well. This could involve feature importance analysis, examining decision boundaries, or comparing with other classical ML techniques.
- Exploration of Multilingual and Larger Transformer Models: Fine-tuning larger, multilingual pre-trained models (e.g., Afro-XLMR-Large, mDeBERTaV3, AfriBERTa-large) on the KannySenti dataset and conducting a more in-depth comparison with the current BERT and RoBERTa results is a key next step.
- Advanced Fine-tuning Techniques: Experimenting with more advanced fine-tuning techniques, such as multi-task learning (e.g., incorporating related tasks like topic classification), or utilizing techniques specifically designed for low-resource scenarios, could lead to improved deep learning model performance.